%% file: main.tex
\documentclass[10pt,twocolumn,letterpaper]{article}

\usepackage{wacv}              % To produce the CAMERA-READY version

\usepackage[accsupp]{axessibility}
\usepackage{multirow}

\definecolor{wacvblue}{rgb}{0.21,0.49,0.74}
\usepackage[pagebackref,breaklinks,colorlinks,allcolors=wacvblue]{hyperref}

\usepackage[table]{xcolor}          % colortbl 을 함께 불러온다
\definecolor{selcell}{HTML}{FFF3B0}  % 밝은 노랑

\def\wacvPaperID{68} % *** Enter the WACV Paper ID here
\def\confName{WACV}
\def\confYear{2027}

\title{TOGBench: Benchmarking and Improving Tiny-Object Generation Fidelity in Aerial Imagery}

\author{Chanhwi Jeong$^*$ \hspace{0.5cm} Dongmin You$^*$ \hspace{0.5cm} Seungwook Kim\vspace{0.2cm}\\
GenGenAI, South Korea
}

\begin{document}
\maketitle
% \twocolumn[{
% \renewcommand\twocolumn[1][]{#1}%
% \maketitle
% \vspace{-2.0mm}
% \includegraphics[width=1.0\linewidth]{figures/tog_teaser.pdf}
% % \vspace{-3.mm}
% \captionof{figure}{\textbf{Teaser illustration of TOGBench, and our proposed techniques for improve small-object generation in aerial images.}
% We show that existing evaluation metrics (CLIPScore~\cite{hessel2021clipscore}, PickScore~\cite{kirstain2023pickapic}) cannot fully reflect the fidelity of generated tiny objects. 
% We show that applying our two proposed techniques, Layout-weighted loss and Condition-Layout Attention Gain, visually improves tiny object generation quality.
% We finally show that our proposed TOGBench successfully reflects the improvement in tiny object generation.
% } 
% \vspace{+11.0mm}
% \label{fig:teaser}
% }]

\input{sec/0_abstract} 
\input{sec/1_intro}
\input{sec/2_relatedwork}
\input{sec/3_preliminary}
\input{sec/4_togbench}
\input{sec/5_method}
\input{sec/6_experiments}
\input{sec/7_conclusion}

\clearpage
\clearpage

\noindent
\textbf{Acknowledgement.} 
This research was supported by the MSIT (Ministry of Science and ICT), Korea, under the Defense ICT Innovation Technology program (RS-2026-25539289) supervised by the IITP (Institute of Information \& Communications Technology Planning \& Evaluation).

{
    \small
    \bibliographystyle{ieeenat_fullname}
    \bibliography{main}
}

\clearpage
\input{supp}

\end{document}

%% file: sec/0_abstract.tex
\begin{abstract}
Advances in deep generative models now enable the synthesis of realistic imagery. 
However, aerial imagery is dominated by tiny objects, such as vehicles, ships, and aircraft, which occupy only a few pixels owing to the overhead, wide-area nature of aerial imaging. 
% Existing generative models are markedly weaker at producing these tiny objects than at producing larger ones. 
Existing generative models struggle substantially more with tiny objects than with larger ones.
% In aerial imagery an object is commonly only 8 to 16 pixels across, which leaves little room to render its structure correctly. 
This limitation has been overlooked, as prior work on generative evaluation emphasizes prompt following or human preference rather than whether small objects are rendered faithfully.
To address this gap, we introduce TOGBench, a quantitative, object-level benchmark for evaluating tiny-object generation fidelity in aerial imagery.
% TOGBench compares generated objects to their real counterparts on a per-class basis, along five interpretable axes, and aggregates them into Detectability and Appearance scores.
% A controlled degradation study confirms the benchmark is well validated: TOGBench scores track the severity of progressively applied degradations, so they correctly reflect the quality of generated small objects.
We further propose two simple train-/inference-time techniques that improve the ability of layout-to-image generators to render small objects, raising their performance on TOGBench.
\end{abstract}

%% file: sec/1_intro.tex
\section{Introduction}
\label{sec:intro}
Deep generative models have advanced rapidly in image synthesis \cite{goodfellow2014gan,kingma2014vae,oord2017vqvae,karras2018progan,brock2019biggan,karras2019stylegan,karras2020stylegan2,esser2021vqgan,ramesh2021dalle,yu2022parti,chang2022maskgit,esser2024sd3,tian2024var,blackforest2024flux}. Diffusion models produce
high-fidelity scenes \cite{ho2020ddpm,song2021ddim,song2021scoresde,nichol2021improvedddpm,dhariwal2021beatgans,ho2022cfg,nichol2022glide,rombach2022ldm,saharia2022imagen,ramesh2022dalle2,karras2022edm,podell2024sdxl}, and spatial-control methods extend them to layout-conditioned generation, in
which the positions and classes of objects are specified in advance, commonly as bounding boxes \cite{zhang2023controlnet,li2023gligen,mou2024t2iadapter,zheng2023layoutdiffusion,yang2023reco,avrahami2023spatext,xie2023boxdiff,zhao2023unicontrolnet,huang2023composer,wang2024instancediffusion,zhou2024migc,feng2024ranni,phung2024attentionrefocusing}. These
capabilities have also been extended to aerial imagery \cite{khanna2024diffusionsat,tang2024crsdiff,yu2024metaearth,sastry2024geosynth,zheng2024changen2,liu2025text2earth,zhang2024ccdiff}. Given a layout, a controllable generator can synthesize an
aerial scene with objects such as vehicles, ships, and aircraft at prescribed locations. Such synthetic
imagery is appealing for remote sensing, where annotated data is scarce and costly to acquire \cite{xia2018dota,lam2018xview,zamir2019isaid,li2020dior,wang2021aitod,sun2021fair1m,manas2021seco}. In
particular, it offers a way to augment training data for aerial object detection \cite{chen2024geodiffusion,zhang2023diffusionengine,zhao2023xpaste,ghiasi2021copypaste,zhu2024odgen,tang2025aerogen,ye2025objectfidelity,tang2025terragen}.

These generators face a basic difficulty when generating tiny objects.
In particular, aerial imagery is dominated by tiny objects, often only 8 to 16 pixels across, because the sensor observes a wide area from above~\cite{wang2021aitod,xu2022detecting}.
Generating such objects is far harder than generating large, prominent ones.
The model must place each object correctly and render its shape and scale within only a few pixels.
Current models degrade sharply in this regime~\cite{ye2025objectfidelity,tang2025aerogen,chen2024geodiffusion}, often omitting or distorting the requested objects.
This failure is largely invisible to existing evaluation.
FID and related metrics assess the realism of an entire image distribution and are insensitive to regions that occupy little of the image~\cite{heusel2017gans,binkowski2018demystifying,jayasumana2024rethinking}.
Prompt-following metrics~\cite{hessel2021clipscore,huang2023t2icompbench,ghosh2023geneval} and human-preference metrics~\cite{xu2023imagereward,wu2023human,kirstain2023pickapic} concern semantic alignment rather than geometric fidelity.
The community therefore lacks a way to measure how well tiny objects are generated, and consequently a way to improve it.
This gap matters because aerial applications depend on exactly these tiny objects~\cite{cheng2023towards,wang2023small}.
Synthetic imagery whose tiny objects are malformed or mislocated has little value for training a detector, and can even degrade it~\cite{vanherle2022analysis,chen2024geodiffusion}.

To close this gap, we introduce TOGBench, a benchmark for evaluating tiny-object generation fidelity in aerial imagery.
Rather than scoring image realism in aggregate, TOGBench evaluates a generator by comparing its objects to real objects of the same class, across seven object categories.
It decomposes fidelity into five interpretable axes, grouped into two complementary scores: Detectability and Appearance.
% Four axes are derived from object detectors used as a perception proxy~\cite{jocher2026yolo26,hu2025domedetr,huang2024dqdetr}: presence, localization, class, and size.
% These four axes combine through a non-compensatory aggregation into a single Detectability score, so that one severely failing axis cannot be masked by the others.
% The fifth axis, appearance, captures a property that detection cannot: it compares generated and real objects through a distributional feature distance in a satellite-pretrained embedding~\cite{simeoni2025dinov3}, and forms the Appearance score.
Because every axis is measured relative to real objects, a detector's own misses on real objects are discounted, and each axis mainly reflects a generation deficit rather than the measuring instrument, with the same ranking of methods under each detector alone.
% We report Detectability and Appearance separately rather than merging them, since the two capture different and non-substitutable aspects of tiny-object fidelity.
We build TOGBench on the AI-TOD dataset~\cite{wang2021aitod,xu2022detecting}.

We additionally propose two simple techniques to improve tiny-object generation.
% , one applied during training and the other at inference.
Layout-Weighted Loss reweights the training objective to emphasize the small bounding-box regions containing target objects.
% , so that the model spends more of its capacity rendering tiny objects rather than the dominant background.
Condition-Layout Attention Gain amplifies attention from image and text tokens to the spatial condition, with an extra gain on cells holding sub-token objects, which improves the generator's adherence to the specified layout.
Both techniques improve the TOGBench performance of baseline generators.

Our contributions are threefold:

\begin{itemize}
    \item We introduce TOGBench, a benchmark to measure the fidelity of layout-conditioned tiny-object generation in aerial imagery via Detectability and Appearance scores.
    \item We analyze a state-of-the-art controllable generator with TOGBench and validate the benchmark with a controlled degradation study.
    \item We propose two techniques, Layout-Weighted Loss and Condition-Layout Attention Gain, that improve tiny-object generation.
\end{itemize}

%% file: sec/2_relatedwork.tex
\section{Related Work}
\label{sec:relatedwork}

\subsection{Computer Vision for Tiny Objects}
Object detection is the task of localizing and classifying every object in an image~\cite{zou2023survey,liu2020deep,girshick2014rcnn,he2014spp,girshick2015fastrcnn,ren2015fasterrcnn,redmon2016yolo,liu2016ssd,redmon2018yolov3,lin2017focal, tan2020efficientdet,ge2021yolox,wang2023yolov7,carion2020detr,zhu2021deformabledetr,liu2022dabdetr,li2022dndetr,zhang2023dino,zong2023codetr,zhao2024rtdetr}.
Detecting tiny objects is one of its hardest cases, because an object that spans only a few pixels carries little appearance information and is easily lost against the background~\cite{cheng2023towards}.
Detectors evolved along three lines.
Two-stage detectors generate region proposals and classify them~\cite{girshick2014rcnn}, and later work shared computation~\cite{girshick2015fastrcnn,he2014spp} and learned the proposals end-to-end~\cite{ren2015fasterrcnn}.
One-stage detectors predict boxes densely for speed~\cite{redmon2016yolo,liu2016ssd}, and later variants added a focal loss and anchor-free heads~\cite{lin2017focal,tian2019fcos}.
Transformer-based detectors cast detection as set prediction~\cite{carion2020detr}, with deformable and query-based variants that improve small-object recall~\cite{zhu2021deformabledetr,zhang2023dino}.
Small objects remained hard throughout, which motivated feature pyramids~\cite{lin2017fpn} and high-resolution or context-aware backbones~\cite{wang2021hrnet,li2019trident}.
The tiny regime, where objects span only a few pixels, is studied through dedicated datasets, including the aerial benchmarks DOTA~\cite{xia2018dota}, VisDrone~\cite{zhu2021visdrone}, and AI-TOD~\cite{wang2021aitod} with its refinement AI-TOD-v2~\cite{xu2022detecting}.
% , and the pedestrian benchmark TinyPerson~\cite{yu2020tinyperson}.
% Measurement is also fragile here, because intersection-over-union is unstable for tiny boxes~\cite{xu2022detecting}, which motivated scale-robust criteria such as the Normalized Wasserstein Distance~\cite{wang2021nwd} and receptive-field-based assignment~\cite{xu2022rfla}.

\subsection{Controllable Image Generation}
Controllable image generation synthesizes images that follow user-specified content and structure~\cite{kim2024correspondentdream,kim2024mvprompt,kim2025rapidmv,kim2025freeaction}.
% The field moved from Generative Adversarial Networks (GANs)~\cite{goodfellow2014gan} to diffusion models.
% GANs introduced adversarial synthesis~\cite{goodfellow2014gan}, and conditional~\cite{mirza2014cgan,isola2017pix2pix} and style-based~\cite{karras2019stylegan,karras2020stylegan2} variants added control and improved realism.
Text-to-image generation first used adversarial models~\cite{reed2016text2img}, later refined with attention and stagewise generation~\cite{xu2018attngan,zhang2017stackgan}.
Denoising diffusion models offered a stable alternative with strong quality~\cite{ho2020ddpm,dhariwal2021beatgans}, and classifier-free guidance improved prompt adherence~\cite{ho2022cfg}.
Latent diffusion made large-scale text-to-image generation practical~\cite{rombach2022ldm}, and later systems improved resolution and fidelity~\cite{podell2024sdxl,esser2024sd3,kim2026solace}.
Such models have also been applied to aerial and remote-sensing imagery for simulation and augmentation~\cite{khanna2024diffusionsat,tang2024crsdiff,tang2025aerogen}.
Text alone gives coarse spatial control, which motivated structural conditioning.
ControlNet~\cite{zhang2023controlnet} and T2I-Adapter~\cite{mou2024t2iadapter} condition on dense maps such as edges or depth, and grounded and layout-to-image methods place objects from bounding boxes~\cite{li2023gligen,zheng2023layoutdiffusion,wang2024instancediffusion,zhou2024migc}.
Low-rank~\cite{hu2022lora}, image-prompt~\cite{ye2023ipadapter}, and visual-prompt~\cite{jeong2025test} adapters make such control inexpensive to add, and unified frameworks support many conditions in one model~\cite{qin2023unicontrol,zhao2023unicontrolnet,huang2023composer}.

In this work, we evaluate and improve a representative model of this kind, a diffusion backbone with a spatial adapter conditioned on class-labeled bounding boxes, applied to aerial scenes~\cite{blackforest2024flux,tan2024ominicontrol}.
% Prior control methods are optimized for layout adherence and image-level quality, and their fidelity on objects of only a few pixels is neither measured nor improved.
% We close this gap on both sides.
We measure tiny-object fidelity with a dedicated benchmark, and we introduce two techniques that improve it.
% : a training objective that reweights the small bounding-box regions, and an inference-time amplification of the condition cross-attention that strengthens layout adherence.

\subsection{Evaluating Generated Images}
Evaluating generated images is an active research area with several distinct lines.
Early work used the Inception Score, which rewards quality and diversity through a pretrained classifier~\cite{salimans2016is}.
The Fréchet Inception Distance (FID) became the standard measure of distributional realism~\cite{heusel2017gans}, and was refined by the Kernel Inception Distance~\cite{binkowski2018demystifying}, improved precision and recall~\cite{kynkaanniemi2019precision}, and density and coverage~\cite{naeem2020density}.
Later work replaced Inception features with CLIP or self-supervised ones, as in CLIP-FID~\cite{kynkaanniemi2023clipfid} and DINO feature distances~\cite{stein2023exposing}.
A second line measures image-text alignment. CLIPScore scores similarity without a reference~\cite{hessel2021clipscore}, and compositional benchmarks such as GenEval~\cite{ghosh2023geneval}, T2I-CompBench~\cite{huang2023t2icompbench}, TIFA~\cite{hu2023tifa}, and VQAScore~\cite{lin2024vqascore} test attributes, counting, and spatial relations, and SOEBench~\cite{pan2024smallobject} evaluates text-guided editing of small objects in natural images.
A third line models human preference with reward models such as ImageReward~\cite{xu2023imagereward}, PickScore~\cite{kirstain2023pickapic}, and HPSv2~\cite{wu2023hpsv2}.
A fourth line measures a functional property directly, for example reading rendered text with optical character recognition~\cite{chen2023textdiffuser} or judging synthetic data by downstream performance~\cite{he2023is,vanherle2022analysis,nam2026ccdm}.

In this work, 
% we evaluate generation at the level of individual tiny objects rather than the whole image.
% The measures above capture global realism, semantic alignment, or human preference, and they overlook objects that fill only a small fraction of the image.
we propose an object-level, per-class benchmark that compares generated objects to real ones of the same class along five interpretable axes, and reports two complementary scores: a non-compensatory Detectability score from detection-based measurements, and an Appearance score from a distributional feature distance in a satellite-pretrained embedding~\cite{simeoni2025dinov3}.
% We validate the benchmark with a controlled degradation study, in which both scores track the severity of applied degradations.
% To our knowledge, TOGBench is the first benchmark to measure the fidelity of tiny-object generation.

%% file: sec/3_preliminary.tex
\section{Preliminary: Layout-to-Image
Generation}
\label{sec:prelim}

Diffusion models generate images by gradually denoising a sample drawn from noise~\cite{sohldickstein2015diffusion,ho2020ddpm,song2021scoresde}. Given a clean image $x_0$, a forward process adds noise to produce $x_t = \sqrt{\bar{\alpha}_t}\,x_0 + \sqrt{1-\bar{\alpha}_t}\,\epsilon$ at step $t$, with $\epsilon \sim \mathcal{N}(0, \mathbf{I})$~\cite{ho2020ddpm}. A text-to-image model trains a denoiser $\epsilon_\theta$ to predict this noise,
\begin{equation}
\mathcal{L}_{\mathrm{t2i}} = \mathbb{E}_{x_0,\epsilon,t}\left[\left\lVert \epsilon - \epsilon_\theta(x_t, t, y) \right\rVert_2^2\right].
\label{eq:t2i}
\end{equation}
The prompt $y$ specifies content but not precise spatial arrangement. Two images that satisfy the same prompt can place objects very differently.

Layout-to-image generation adds spatial control by conditioning on a layout. A layout is a set of objects with their positions and classes, most often given as class-labeled bounding boxes~\cite{zheng2023layoutdiffusion,yang2023reco,zhou2024migc}. We write it as $L = \{(b_i, c_i)\}_{i=1}^{N}$, where $b_i$ is a bounding box and $c_i$ is its class. The denoiser then conditions on the layout in addition to the prompt, $\epsilon_\theta(x_t, t, y, L)$, and is trained with
\begin{equation}
\mathcal{L}_{\mathrm{l2i}} = \mathbb{E}_{x_0,\epsilon,t}\left[\left\lVert \epsilon - \epsilon_\theta(x_t, t, y, L) \right\rVert_2^2\right].
\label{eq:l2i}
\end{equation}
The layout $L$ is supplied alongside the prompt, which lets the model control where each object appears. 
Methods differ in how the layout is injected. Some render it as a spatial map and add it through an adapter~\cite{zhang2023controlnet,mou2024t2iadapter}, while others encode each box as a grounding token~\cite{li2023gligen,wang2024instancediffusion}.

We study a representative layout-to-image generator, a pretrained diffusion backbone with a lightweight spatial adapter~\cite{blackforest2024flux,tan2024ominicontrol}. The layout is supplied as a bounding-box mask in which each object is drawn as a filled box colored by its class. 
% Given the mask and a prompt, the model synthesizes a scene that should contain each requested object at its specified location, class, and size. 
The layout therefore acts as an explicit ground truth, which lets us compare every generated object against the real object that the layout prescribes.

%% file: sec/4_togbench.tex
\begin{figure*}[t]
\centering
\includegraphics[width=0.9\linewidth]{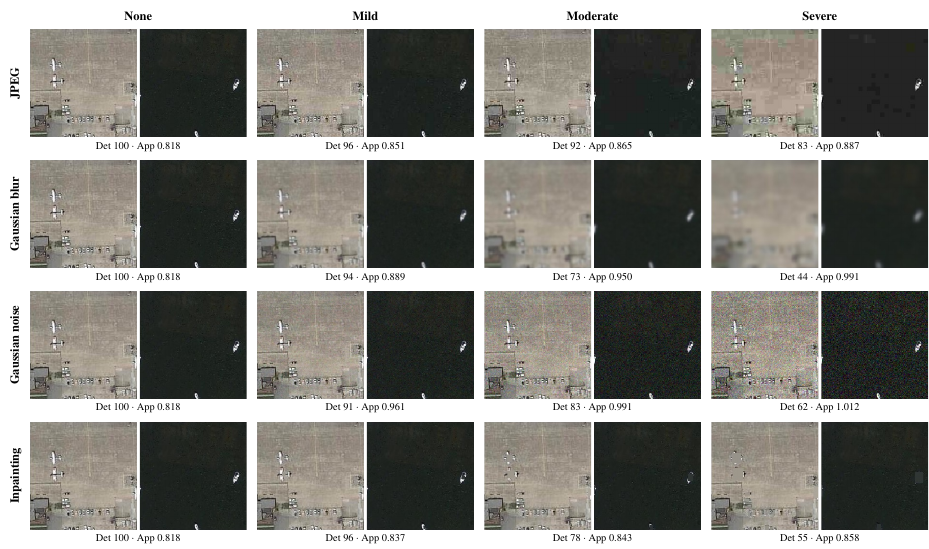}
\vspace{-4mm}
\caption{\textbf{Both TOGBench scores respond to changes in image quality.} Each cell shows two example regions degraded identically, a vehicle scene (left) and a ship scene (right); these regions are cropped only for display.
Below each cell we report Detectability and Appearance measured over a $100$-scene subset degraded to that setting. Both scores successfully track image quality as it degrades.}
\vspace{-5mm}
\label{fig:validation}
\end{figure*}

\section{TOGBench Benchmark}
\label{sec:togbench}

\subsection{Overview}
\label{subsec:overview}
TOGBench measures how faithfully a layout-to-image generator renders tiny objects, following three principles. First, every property is measured relative to real objects of the same class, so the gap between the real and generated sets reduces the bias of the detector or feature extractor. Second, we decompose fidelity into five interpretable axes, one per failure mode: presence, localization, class, size, and appearance. Third, the four detection axes combine non-compensatorily into a Detectability score, and appearance forms a separate Appearance score.

% Given a generator, we synthesize images from the real test layouts and measure the five axes per class against the real objects. Four axes come from object detectors used as a perception proxy and form the Detectability score (\S\ref{subsec:det-axes}); the fifth measures appearance, which detection cannot capture, and forms the Appearance score (\S\ref{subsec:app-shape}). We report the per-axis profile together with the two scores, which are kept separate and never merged into a single number.

\subsection{Detection-Derived Axes}
\label{subsec:det-axes}
Four axes describe how object detectors perceive the generated objects: presence, localization, class, and size. We run three detectors~\cite{jocher2026yolo26,hu2025domedetr,huang2024dqdetr}, each fine-tuned on AI-TOD~\cite{wang2021aitod,xu2022detecting}, on both the real and the generated images, keep every detection with confidence at least $\tau = 0.3$, and compute each axis per class and per image set, pooled over the three detectors. The ensemble reduces the influence of any single detector.
% , and measuring each axis from an explicit box-to-detection matching, rather than from a single summary such as average precision, keeps existence and quality separate.

\smallbreak
\noindent\textbf{Object matching.} Each axis needs to know which detection corresponds to which layout box. Let a box $b$ have center $\mu_b$ and area $A_b$, and a detection $d$ have center $\mu_d$, area $A_d$, and predicted class $\hat{c}_d$. A detection is eligible for $b$ when its center falls within a size-scaled radius,
\begin{equation}
\lVert \mu_d - \mu_b \rVert \le \kappa \sqrt{A_b}, \qquad \kappa = 1,
\end{equation}
so the tolerance grows with object size and does not favor larger boxes. Among all eligible pairs we assign detections to boxes greedily in order of increasing distance, using each detection at most once. Matching is class-agnostic, so placement and class are measured independently. We write $M(b)$ for the detections matched to $b$, call $b$ \emph{found} when $M(b) \neq \emptyset$, and let $B_c$ and $F_c$ denote the boxes and the found boxes of class $c$.

\smallbreak
\noindent\textbf{Presence.} The fraction of layout boxes that are found,
\begin{equation}
\mathrm{Pre}_c = \frac{1}{|B_c|}\sum_{b \in B_c} \mathbb{1}[\,M(b) \neq \emptyset\,],
\end{equation}
i.e. whether an object appears where the layout requests.

\smallbreak
\noindent\textbf{Localization.} For each found box we take its best overlap with a matched detection and average over found boxes,
\begin{equation}
\mathrm{Loc}_c = \frac{1}{|F_c|}\sum_{b \in F_c} \max_{d \in M(b)} \mathrm{IoU}(b, d).
\end{equation}
We report the overlap as intersection-over-union.
% because it is unstable for tiny boxes, the supplementary material also reports the Normalized Wasserstein Distance~\cite{wang2021nwd}.
Localization asks how accurately a found object is placed.

\smallbreak
\noindent\textbf{Class.} Among the detections matched to found boxes, the fraction whose predicted class equals the box class,
\begin{equation}
\mathrm{Cls}_c = \frac{\sum_{b \in F_c}\sum_{d \in M(b)} \mathbb{1}[\,\hat{c}_d = c\,]}{\sum_{b \in F_c} |M(b)|}.
\end{equation}
This isolates recognition from presence and placement.

\smallbreak
\noindent\textbf{Size.} A matched pair $(b,d)$ is mis-sized when its size ratio $\rho = \sqrt{A_d / A_b}$ exceeds $\eta$ or falls below $1/\eta$ ($\eta = 1.3$),
\begin{equation}
\mathrm{Siz}_c = \frac{1}{|\mathcal{M}_c|}\sum_{(b,d) \in \mathcal{M}_c} \mathbb{1}\!\left[\,\rho > \eta \;\lor\; \rho < 1/\eta\,\right],
\end{equation}
$\mathcal{M}_c$ being the set of matched box-detection pairs of class $c$. 
% Size is two-sided; reporting the too-large and too-small cases separately distinguishes oversizing from undersizing.

\smallbreak
\noindent\textbf{From axes to Detectability.} We score each axis against its real anchor rather than in absolute terms. Let $R_{c,a}$ and $G_{c,a}$ be axis $a$ of class $c$ on the real and generated sets; the severity is the clamped relative deficit,
\begin{equation}
s_{c,a} =
\begin{cases}
\big[(R_{c,a} - G_{c,a})/R_{c,a}\big]_0^{\sigma}, & a \in \{\text{pre},\text{loc},\text{cls}\},\\[3pt]
\big[(G_{c,a} - R_{c,a})/(1 - R_{c,a})\big]_0^{\sigma}, & a = \text{size},
\end{cases}
\end{equation}
with $[x]_0^{\sigma} = \min(\max(x,0),\,\sigma)$ and $\sigma = 1.5$, so $s_{c,a} = 0$ means generation matches or beats the real anchor. Per-class Detectability is a non-compensatory power mean over the four axes $A_c$,
\begin{equation}
D_c = 100\left(1 - \Big(\tfrac{1}{|A_c|}\textstyle\sum_{a \in A_c} s_{c,a}^{\,p}\Big)^{1/p}\right), \qquad p = 2,
\label{eq:detectability-class}
\end{equation}
% quality axes (loc, class, size) require >= N_min=50 matched detections per class; presence is always kept.
and the overall score, written $\mathrm{Det}_4$ after the four detection axes, macro-averages the classes the same way,
\begin{equation}
\mathrm{Det}_4 = 100\left(1 - \Big(\tfrac{1}{|C|}\textstyle\sum_{c \in C} (1 - D_c/100)^{p}\Big)^{1/p}\right).
\label{eq:detectability}
\end{equation}
Higher is better ($100$ = as detectable as real); $p = 2$ makes the aggregation non-compensatory, so one weak axis or class cannot be masked.

\subsection{Appearance}
\label{subsec:app-shape}
A detector can find an object without judging whether it looks realistic. The Appearance axis measures this. 

We crop at the layout boxes, with some surrounding context, from the real and the generated images, so an omitted object is scored on the content of its box, and embed each crop with DINOv3 pretrained on satellite imagery~\cite{simeoni2025dinov3}. Each crop is embedded at two small scales ($32$ and $48$ px), and the two features are $L_2$-normalized and concatenated. For a generated crop of class $c$ with feature $g$, we compute the mean distance to its $k = 25$ nearest real crops of the same class in a fixed reference set $\mathcal{R}_c$,
\begin{equation}
a(g) = \frac{1}{k}\sum_{r \in \mathrm{NN}_k(g,\, \mathcal{R}_c)} \lVert g - r \rVert_2 .
\label{eq:appearance-obj}
\end{equation}
The Appearance of a class is the mean of $a(g)$ over its generated crops, and the reported score is the mean over classes. 
% Lower is better. We use the distance to nearest real crops rather than a Fréchet distance, because a Fréchet distance assumes many samples and a Gaussian feature distribution, which do not hold for the few tiny crops in each class. 
% A generator cannot be closer to real than real data itself, so we also report a real-real floor, computed the same way with held-out real crops, and the gap above it.

\subsection{Setup and Validation}
\label{subsec:setup}
\smallbreak
\noindent\textbf{Instruments.} The detection axes use an ensemble of three detectors: a one-stage detector~\cite{jocher2026yolo26}, a query-based transformer detector~\cite{hu2025domedetr}, and a detector specialized for tiny objects~\cite{huang2024dqdetr}. All three are fine-tuned on the target aerial domain. The appearance axis uses features from a self-supervised transformer pretrained on satellite imagery~\cite{simeoni2025dinov3}. 
% Combining several detectors of different designs, rather than relying on one, keeps the detection axes from reflecting the bias of any single model.

\smallbreak
\noindent\textbf{Data.} We build TOGBench on AI-TOD~\cite{wang2021aitod,xu2022detecting}, an aerial dataset of tiny objects, and use seven classes: airplane, bridge, storage-tank, ship, swimming-pool, vehicle, and person. The benchmark is a fixed set of $500$ real scenes, selected from the AI-TOD-v2 test split containing $4{,}459$ annotated objects in total.
% : airplane $525$, bridge $557$, storage-tank $581$, ship $635$, swimming-pool $249$, vehicle $1{,}266$, and person $646$. 
The objects are small, with a median size of about $12$ pixels (measured as $\sqrt{\text{area}}$) and most objects between $8$ and $22$ pixels. AI-TOD does not include text prompts, so we generate one per scene by captioning the real image with a vision-language model, Qwen3-VL-32B-Instruct~\cite{bai2025qwen3vl}. We then give the generator the scene's real layout with this prompt, produce one image at $800 \times 800$ resolution, and measure every axis per class against the real objects. 
% The three detectors (YOLO26x, Dome-DETR, and DQ-DETR) are each fine-tuned on AI-TOD-v2 and run at the native $800 \times 800$ resolution with a confidence threshold of $\tau = 0.3$. 
The appearance axis compares against a fixed reference bank of $24{,}933$ real object crops from the same dataset, including $2{,}345$ from the benchmark scenes.
% ; the real-real floor, the distance of held-out real crops to this bank, averages $0.825$ across classes and ranges from $0.80$ (ship) to $0.86$ (vehicle).

\smallbreak
\noindent\textbf{Validation.} We validate TOGBench with controlled degradations of two kinds. First, we degrade real scenes of known quality with several corruptions, including JPEG compression, Gaussian blur, Gaussian noise, and object inpainting, at graded severity. Under every corruption, as severity increases both scores worsen: Detectability falls and Appearance distance rises (Figure~\ref{fig:validation}). Second, we perturb the layout that conditions generation: dropping boxes raises the presence severity and leaves the other axes near the baseline, rescaling boxes raises size and localization, since a resized object overlaps its box less, and relabeling raises class only modestly. See the supplementary material for details. 

%% file: sec/5_method.tex
\begin{figure}[t]
\centering
\includegraphics[width=\linewidth]{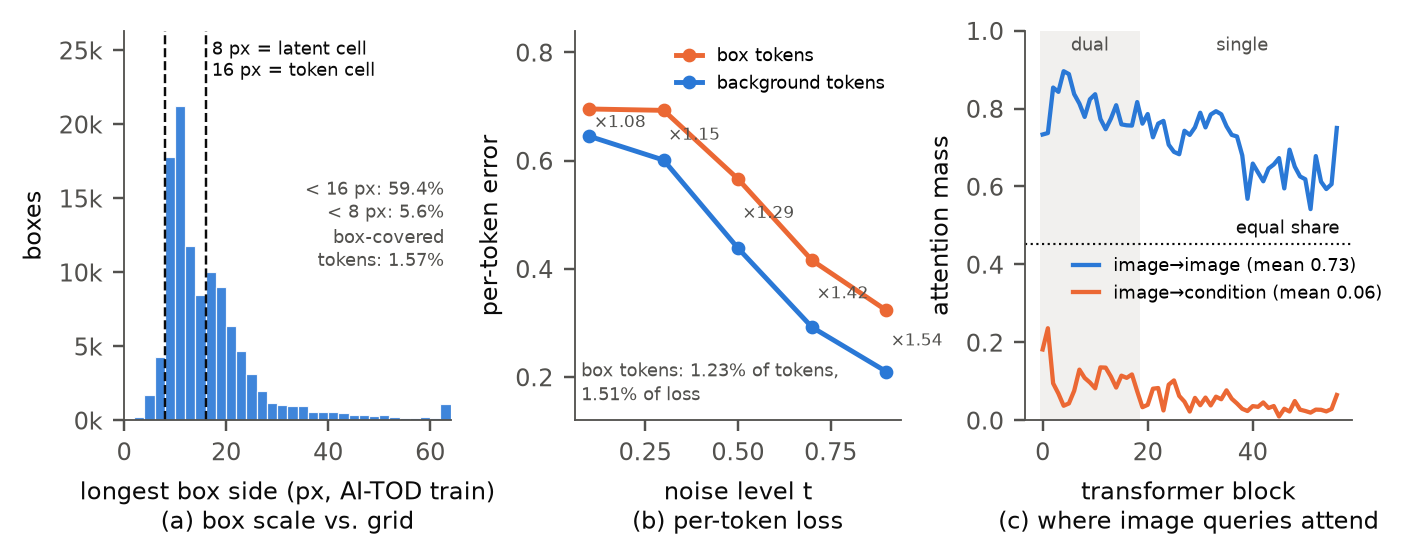}
\vspace{-4mm}
\caption{Where the layout signal is lost (baseline generator). (a) AI-TOD box sizes vs.\ the latent-cell ($8$,px) and token-cell ($16$,px) scales: most boxes are smaller than one token cell and cover $\sim$$1.5\%$ of latent locations. (b) Per-location denoising error, box vs.\ background: boxes are harder at every noise level, yet the uniform loss weights them only by their token fraction. (c) Image-query attention to image vs.\ condition keys per block: despite equal key counts, the condition receives only a small share, less still in single-stream blocks.}
\label{fig:diagnosis}
\vspace{-2mm}
\end{figure}

\begin{figure}[t]
\centering
\includegraphics[width=\linewidth]{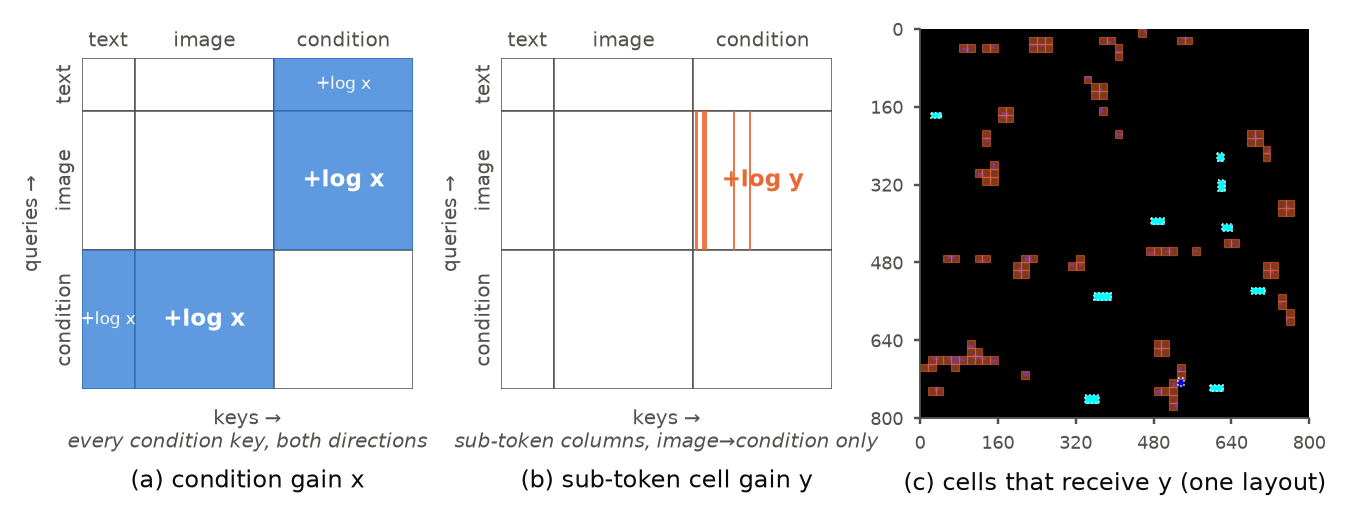}
\vspace{-4mm}
\caption{Two attention gains as edits to the joint attention mask $[\text{text},|,\text{image},|,\text{condition}]$. (a) Condition gain: add $\log x$ to all condition keys for image/text queries, and symmetrically to image/text keys for condition queries. (b) Sub-token-cell gain: add $\log y$ to the columns of cells holding a box smaller than a token cell, for image queries only. (c) Cells receiving $y$ in one benchmark layout (orange), over the layout mask; boxes at or above cell size (white outline) get no gain.}
\label{fig:method}
\vspace{-2mm}
\end{figure}

\begin{figure}[t]
\centering
\includegraphics[width=\linewidth]{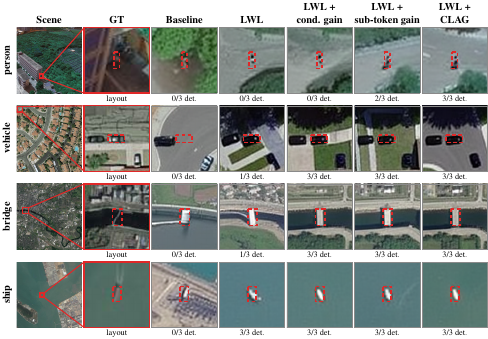}
\vspace{-5mm}
\caption{Qualitative effect of each component, added cumulatively on four benchmark layouts. The dashed box is the layout, and ``$n$/3 det.'' counts the detectors that find the object. LWL: Layout-Weighted Loss.}
\label{fig:cumeff}
\vspace{-4mm}

\end{figure}

\section{Improving Tiny-Object Generation}
\label{sec:method}

Our benchmark identifies presence and class, the two axes that ask whether the layout was followed at all, as the dominant failures of tiny-object generation, as we show in Section~\ref{subsec:main-results} (Table~\ref{tab:tog-main}).
% The layout is supplied, so the failure is the model not following it. 
We trace this to three places where the layout signal is diluted, measure each, and propose targeted remedies.
The first, Layout-Weighted Loss, changes training by reweighting the denoising loss to raise the share of box-covered locations in the objective (Section~\ref{subsec:layerweight}). The latter two are training-free and together form Condition-Layout Attention Gain (CLAG), which edits the attention of the spatial adapter at inference through a global gain on the condition stream and a gain restricted to the cells that hold sub-token objects (Section~\ref{subsec:condstream}).

\subsection{Where the Layout Signal Is Lost}\label{subsec:siglost}
\smallbreak\noindent\textbf{Training objective.} The layout-to-image objective in Equation~\eqref{eq:l2i} averages the denoising error uniformly over latent locations. Let $\mathcal{B}$ be the set of latent locations covered by any box in $L$ and $f = |\mathcal{B}| / |U|$ its fraction of all locations $U$. Under a uniform average the objective allocates them only the fraction $f$ of its weight, regardless of how hard they are to predict. In AI-TOD the layouts are sparse: the covered fraction is about 1.5\% of the latent grid (Figure~\ref{fig:diagnosis}a). We measure the per-location error of the baseline generator separately for covered and background locations (Figure~\ref{fig:diagnosis}b). Covered locations are harder at every noise level, and increasingly so as noise grows, yet their weight in the objective stays at the token fraction. The objective spends the least effort on the locations that are the hardest.
\smallbreak\noindent\textbf{Condition attention.} The generator we study injects the layout as a stream of condition tokens concatenated with the text and image tokens, and all three streams interact through a single joint self-attention; there is no dedicated cross-attention for the condition~\cite{tan2024ominicontrol}. Image tokens therefore divide their attention among text, image, and condition keys, and the condition competes with the image's own self-attention. The image and condition streams have the same number of tokens, so under an even split each would receive the same mass (Figure~\ref{fig:diagnosis}c). 
% Figure~\ref{fig:diagnosis}c shows the attention mass that image queries assign to each stream, per block, in the baseline. 
The condition receives a small fraction of what the image assigns to itself, and the fraction falls further in the later single-stream blocks. 
% The stream that specifies where every object goes is read far less than the image reads itself.

\smallbreak\noindent\textbf{Token resolution.} Each (latent) condition token covers a $16{\times}16$ pixel cell. A box whose longer side is under 16 px cannot own a token; it shares a cell with its surroundings, and the cell's key carries a mixture of object and background. In AI-TOD this is the majority case: over half of the training boxes are smaller than a token cell (Figure~\ref{fig:diagnosis}a). A global gain on the condition stream does not change this, because it scales every condition key alike and leaves the object-to-background ratio inside a cell untouched. 
% This dilution is therefore separate from the previous one, and it calls for a gain that acts on those cells specifically.

\subsection{Layout-Weighted Loss}\label{subsec:layerweight}
\smallbreak\noindent\textbf{Motivation.} Section~\ref{subsec:siglost} diagnosed that box-covered locations as a whole are under-weighted, not that small boxes are under-weighted relative to large ones. We therefore raise the weight of every covered location by the same factor and let the normalization redistribute the objective.

\smallbreak\noindent\textbf{Method.} For a latent location $u$ we set
\begin{equation}
w(u) = \begin{cases} W & u \in \mathcal{B} \\ 1 & \text{otherwise,}\end{cases} 
\label{eq:box-weight}
\end{equation}
where $\mathcal{B}$ is the set of locations intersected by any box of $L$ on the latent grid. The weighted objective is
\begin{equation}
\mathcal{L}_{\mathrm{rw}} = \mathbb{E}_{x_0,\epsilon,t}\left[\frac{\sum_u w(u)\,\|\epsilon_u - \epsilon_\theta(x_t,t,y,L)_u\|_2^2}{\sum_u w(u)}\right].
\label{eq:weighted-loss}
\end{equation}
The denominator keeps Equation~\eqref{eq:weighted-loss} a proper weighted average, and $W{=}1$ recovers Equation~\eqref{eq:l2i} exactly. 
% Under Equation~\eqref{eq:weighted-loss} 
The share of the objective that covered locations receive is
\begin{equation}
s(W) = \frac{W f}{W f + (1 - f)},
\label{eq:obj-weighted-loss}
\end{equation}
which rises from $f$ at $W{=}1$ and saturates as $W$ grows; for the sparse layouts of AI-TOD, a moderate $W$ multiplies the share several-fold while leaving the background dominant. 
% We treat $W$ as the single hyperparameter of this component and study it in Section~\ref{subsec:ablations}.

\subsection{Condition-Layout Attention Gain}\label{subsec:condstream}
Both remedies below edit the attention logits of the spatial adapter at inference and require no training. Let $\alpha_{ij}$ be the attention from query $i$ to key $j$, and let $\mathcal{C}$ denote the set of condition keys. Adding a constant $\log g_j$ to the logit of key $j$ before the softmax gives
\begin{equation}
\alpha_{ij} = \frac{g_j \exp(q_i^\top k_j/\sqrt{d})}{\sum_{j'} g_{j'} \exp(q_i^\top k_{j'}/\sqrt{d})},
\label{eq:att-condstream}
\end{equation}
so the mass on key $j$ is multiplied by $g_j$ and renormalized. The edit changes how much each key is read; it does not move a token, add one, or alter the value it carries. We use two instances of Equation~\eqref{eq:att-condstream}.
\smallbreak\noindent\textbf{Condition gain.} To counter the under-reading in Figure~\ref{fig:diagnosis}c we set $g_j = x$ for every $j \in \mathcal{C}$ and $g_j = 1$ otherwise, with $x > 1$ (Figure~\ref{fig:method}a). The image and text tokens then read the condition stream $x$ times more, relative to the other streams, at every block. We apply the same gain in the reverse direction as well, from condition queries to image and text keys, because the condition tokens are updated through this attention. Scaling both directions amplifies the condition stream as a whole, rather than only its readout.
% following the condition-scale mechanism of the adapter we build on~\cite{tan2024ominicontrol}. The reverse gain matters
% because the condition tokens are themselves updated through this attention. 
By Equation~\eqref{eq:att-condstream}, $x$ is exactly the odds multiplier on condition keys, and the supplementary verifies that the measured odds track $x$.

\smallbreak\noindent\textbf{Sub-token cell gain.} The attention mask is a $[q, k]$ matrix, and each column of its condition block corresponds to one condition token, i.e., one $16{\times}16$ cell of the layout mask. The gain can therefore differ per cell. Let $\mathcal{S} \subset \mathcal{C}$ be the cells that contain a box whose longer side is under 16 px. We set $g_j = y$ for $j \in \mathcal{S}$ and $g_j = 1$ otherwise, and apply it only from image queries to condition keys (Figure~\ref{fig:method}b). The threshold is not tuned; it is the cell size of the token grid, the scale at which Section~\ref{subsec:siglost} locates this dilution. Unlike the condition gain, this edit is one-directional: its purpose is to change what the image reads from those cells, not how those few condition tokens update themselves. It is also sparse. In a typical AI-TOD layout it touches under one percent of the cells (Figure~\ref{fig:method}c), so it moves little attention mass overall and, as we show in Section~\ref{subsec:ablations}, has a correspondingly small cost in appearance. The two gains compose additively in the logit, so a sub-token cell receives $\log x + \log y$ while every other condition cell receives $\log x$.

\subsection{How the Three Interact}\label{subsec:threeinteract}
% The three remedies address the same symptom through different channels, and they are not independent. 
The two attention gains share a budget. A sub-token cell is amplified by the product $xy$, so raising one lowers the useful range of the other.
Loss reweighting, in turn, trains the model to follow the layout more closely on its own, which reduces how much gain the attention edits should add at inference. 
We therefore expect the optimal $x$ to fall for a reweighted model, and the optimal $y$ to fall as $x$ rises. Section~\ref{subsec:ablations} tests both predictions with a grid over $(x, y)$ on the baseline and on a reweighted model. Figure~\ref{fig:cumeff} shows the cumulative qualitative effect. Reweighting recovers objects the baseline omits, the condition gain sharpens their placement, and the sub-token gain recovers the smallest ones.

%% file: sec/6_experiments.tex
\begin{figure}[t]
\centering
\includegraphics[width=\linewidth]{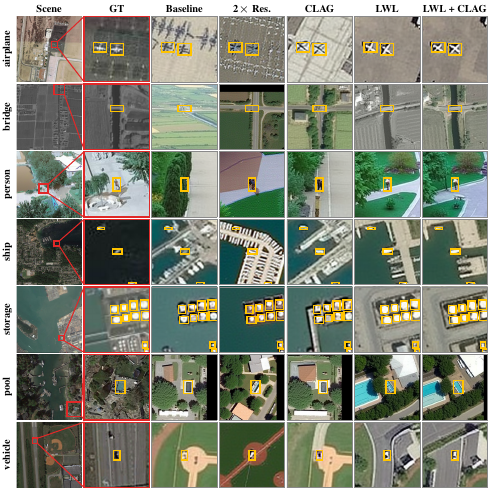}
\vspace{-4mm}
\caption{Qualitative comparison on benchmark layouts. Magnified crops highlight the target tiny objects (yellow boxes). Our CLAG and Layout-Weighted Loss (LWL) improve tiny-object fidelity over the baseline and \(2\times\) inference resolution, with their combination producing the most consistent results.}
\label{fig:qualitative}
\vspace{-4mm}

\end{figure}

\section{Experiments}
\label{sec:exp}
\subsection{Experimental Setup}
\label{subsec:exp-setup}

\smallbreak\noindent\textbf{Generator.} The baseline generator is a spatial-control LoRA adapter (OminiControl~\cite{tan2024ominicontrol}) on FLUX.1-dev~\cite{blackforest2024flux}, which maps a layout of class-labeled bounding boxes to an image. The detectors and the appearance backbone are the instruments described in Section~\ref{subsec:setup}.

\smallbreak\noindent\textbf{Metric.} We report the two TOGBench scores, Detectability (higher is better) and Appearance (a distance to real objects, lower is better), as macro scores, together with the per-axis detection severities.

\smallbreak\noindent\textbf{Protocol.} All methods generate from the same real layouts, and the inference-time method reuses the baseline seed, so every comparison is paired. We report the mean over five random seeds and treat a change as meaningful only if it exceeds the seed-control noise floor established in Section~\ref{subsec:ablations}.

\begin{table*}[t]
\centering
\footnotesize
\caption{\textbf{Main results on TOGBench.} Relative inference time denotes the per-image generation time
normalized to the default configuration at \(1.00\times\).
\textbf{Bold} denotes the best value and \underline{underline} the second best. Values are mean ± standard deviation over five generation seeds.}
\label{tab:tog-main}
\setlength{\tabcolsep}{2.6pt}
\renewcommand{\arraystretch}{1.3}
\newcommand{\gpx}[1]{{\scriptsize #1}}
\newcommand{\sdx}[1]{{\scriptsize$\pm$#1}}
\newcommand{\lrule}{\cmidrule[0.03em](lr){1-10}}
\resizebox{\textwidth}{!}{%
\begin{tabular}{@{}l | *{4}{c} | c | c | *{2}{c} | c@{}}
\toprule
Method & \multicolumn{5}{c|}{Detectability} & Appearance &
  \multicolumn{2}{c|}{Scene-level text--image metrics} & Efficiency \\
\cmidrule(lr){2-6}\cmidrule(lr){7-7}\cmidrule(lr){8-9}\cmidrule(l){10-10}
& Pres.\,$\downarrow$ & Loc.\,$\downarrow$ & Class\,$\downarrow$ &
  Size\,$\downarrow$ & $\mathrm{Det}_4$\,$\uparrow$ & Dist.\,$\downarrow$ &
  CLIPScore\,$\uparrow$ \gpx{(ViT-L/14)} & PickScore\,$\uparrow$ &
  Rel.\ time\,$\downarrow$ \\
\midrule
FLUX.1-dev \gpx{(text-only)}
& 0.969\,\sdx{0.001} & 0.775\,\sdx{0.033} & 0.672\,\sdx{0.026} & 0.730\,\sdx{0.007}
& 7.620\,\sdx{0.420} & 1.005\,\sdx{0.003} & \textbf{0.2604}\,\sdx{0.0008} & \textbf{23.151}\,\sdx{0.039}
& $0.44\times$ \\
\lrule
\multicolumn{10}{@{}l}{FLUX.1-dev $+$ layout} \\
\quad guidance scale $=1.0$
& 0.723\,\sdx{0.017} & 0.198\,\sdx{0.015} & 0.225\,\sdx{0.020} & 0.186\,\sdx{0.022}
& 59.75\,\sdx{0.647} & 0.931\,\sdx{0.002} & 0.2437\,\sdx{0.0008} & 21.263\,\sdx{0.026}
& $1.00\times$ \\
\quad guidance scale $=3.5$ \gpx{(default)}
& 0.356\,\sdx{0.008} & 0.172\,\sdx{0.008} & 0.334\,\sdx{0.018} & 0.203\,\sdx{0.006}
& 72.19\,\sdx{0.269} & \underline{0.886}\,\sdx{0.001} & 0.2592\,\sdx{0.0004} & \underline{22.174}\,\sdx{0.002}
& $1.00\times$ \\
\quad guidance scale $=6.0$
& 0.372\,\sdx{0.013} & 0.180\,\sdx{0.007} & 0.354\,\sdx{0.011} & 0.218\,\sdx{0.009}
& 70.66\,\sdx{0.871} & 0.894\,\sdx{0.001} & 0.2577\,\sdx{0.0004} & 22.074\,\sdx{0.007}
& $1.00\times$ \\
\lrule
\quad $2\times$ inference resolution
& 0.315\,\sdx{0.009} & 0.145\,\sdx{0.007} & 0.296\,\sdx{0.007} & 0.181\,\sdx{0.011}
& 75.43\,\sdx{0.640} & 0.902\,\sdx{0.001} & 0.2444\,\sdx{0.0001} & 21.169\,\sdx{0.021}
& $4.92\times$ \\
\midrule
Ours \gpx{(CLAG)}
& 0.304\,\sdx{0.009} & \underline{0.119}\,\sdx{0.009} & 0.174\,\sdx{0.022} & 0.144\,\sdx{0.008}
& 80.14\,\sdx{1.000} & \textbf{0.885}\,\sdx{0.001} & 0.2528\,\sdx{0.0005} & 21.732\,\sdx{0.003}
& $1.00\times$ \\
Ours \gpx{(Layout-Weighted Loss)}
& \underline{0.188}\,\sdx{0.016} & 0.135\,\sdx{0.002} & \underline{0.089}\,\sdx{0.002} & \underline{0.141}\,\sdx{0.007}
& \underline{81.14}\,\sdx{0.630} & 0.887\,\sdx{0.002} & \underline{0.2595}\,\sdx{0.0006} & 22.169\,\sdx{0.039}
& $1.00\times$ \\
Ours \gpx{(CLAG + Layout-Weighted Loss)}
& \textbf{0.146}\,\sdx{0.018} & \textbf{0.092}\,\sdx{0.004} & \textbf{0.075}\,\sdx{0.016} & \textbf{0.104}\,\sdx{0.005}
& \textbf{85.34}\,\sdx{0.664} & 0.887\,\sdx{0.002} & 0.2539\,\sdx{0.0005} & 21.857\,\sdx{0.016}
& $1.00\times$ \\
\bottomrule
\end{tabular}}
\end{table*}

\begin{table*}[t]
\centering\small
\caption{\textbf{Attention gains on two checkpoints.} $\mathrm{Det}_4$ (higher is better) and Appearance (lower is better) over the condition gain $x$ (rows) and the sub-token cell gain $y$ (columns), on the baseline checkpoint and on the loss-reweighted checkpoint ($W{=}3$). $\dagger$ marks the checkpoint without edits. \textbf{Bold} is the highest $\mathrm{Det}_4$ in each grid. The \colorbox{selcell}{shaded} cell is the one carried to Table~\ref{tab:tog-main} (Section~\ref{subsec:ablations}). Single seed per cell.}
\label{tab:grid}
\setlength{\tabcolsep}{4.5pt}
\begin{tabular}{@{}l ccccc | ccccc@{}}
\toprule
 & \multicolumn{5}{c|}{Baseline checkpoint} & \multicolumn{5}{c@{}}{Reweighted checkpoint ($W{=}3$)} \\
$x \,\backslash\, y$ & 1.0 & 1.2 & 1.5 & 1.8 & 2.0 & 1.0 & 1.2 & 1.5 & 1.8 & 2.0 \\
\midrule
\multicolumn{11}{@{}l}{\emph{$\mathrm{Det}_4$} $\uparrow$} \\
  1 & 72.52$^{\dagger}$ & 73.34 & 73.69 & 74.12 & 74.68 & 81.88$^{\dagger}$ & 82.47 & 82.77 & 83.30 & 83.55 \\
  1.2 & 76.43 & 76.74 & 76.35 & 77.77 & 77.47 & 84.62 & 84.98 & 85.02 & \colorbox{selcell}{85.48} & 85.43 \\
  1.5 & 80.89 & \colorbox{selcell}{80.96} & 80.47 & 81.08 & 80.81 & 86.04 & \textbf{86.11} & 86.06 & 86.04 & 85.80 \\
  1.8 & 81.73 & \textbf{82.03} & 81.71 & 81.53 & 80.98 & 85.47 & 84.89 & 84.62 & 83.94 & 83.28 \\
  2 & 81.46 & 81.17 & 80.70 & 80.33 & 79.78 & 84.25 & 83.69 & 82.90 & 82.94 & 82.27 \\
\midrule
\multicolumn{11}{@{}l}{\emph{Appearance} $\downarrow$} \\
  1 & 0.8854$^{\dagger}$ & 0.8860 & 0.8853 & 0.8854 & 0.8854 & 0.8861$^{\dagger}$ & 0.8855 & 0.8857 & 0.8862 & 0.8864 \\
  1.2 & 0.8842 & 0.8829 & 0.8836 & 0.8834 & 0.8837 & 0.8879 & 0.8876 & 0.8886 & 0.8877 & 0.8876 \\
  1.5 & 0.8846 & 0.8838 & 0.8836 & 0.8853 & 0.8851 & 0.8896 & 0.8892 & 0.8895 & 0.8900 & 0.8912 \\
  1.8 & 0.8857 & 0.8865 & 0.8880 & 0.8893 & 0.8907 & 0.8917 & 0.8918 & 0.8928 & 0.8924 & 0.8938 \\
  2 & 0.8881 & 0.8897 & 0.8920 & 0.8937 & 0.8949 & 0.8949 & 0.8945 & 0.8964 & 0.8972 & 0.8978 \\
\bottomrule
\end{tabular}
\end{table*}

% \begin{table}[t]
% \centering
% \footnotesize
% \caption{Benchmark validation, external validity by axis. Each row is the Spearman correlation between a per-axis severity and downstream real-test mean average precision, across the $15$ train-on-synthetic conditions at the operating epoch. A more negative value means a higher severity predicts a larger loss of downstream accuracy. The geometric axes are the most predictive; the class axis is not.}
% \label{tab:tstr-axis}
% \begin{tabular}{lc}
% \toprule
% Axis & Spearman with downstream mAP \\
% \midrule
% Localization & $-0.86$ \\
% Size & $-0.84$ \\
% Count & $-0.70$ \\
% Presence & $-0.65$ \\
% Over-generation & $-0.44$ \\
% Class & $-0.04$ \\
% \midrule
% Overall (Det-Fid$_6$) & $+0.76$ \\
% \bottomrule
% \end{tabular}
% \end{table}

\subsection{Main Results}
\label{subsec:main-results}
% SCAFFOLD -- PENDING method runs (fork 3834f6d8). Baseline and Real rows are real; all method and baseline-trick cells are placeholders.

Table~\ref{tab:tog-main} evaluates the baseline generator, each remedy of Section~\ref{sec:method}, and their combination in the 500 fixed layouts. We read the table along the split that Section~\ref{sec:method} motivates. Presence and class measure whether the layout was followed at all. Localization and size measure how precisely it was followed. Detectability aggregates the four; Appearance is reported beside it and is not averaged with it. Figure~\ref{fig:qualitative}
provides qualitative comparisons, showing the corresponding improvements in tiny-object fidelity.

\smallbreak\noindent\textbf{Baseline.} Presence and class are the baseline's two worst axes. Table~\ref{tab:tog-main} also evaluates two common remedies for weak conditioning. Raising the classifier-free guidance scale does not raise Detectability (72.19 to 70.66 at guidance 6.0). Generating at twice the resolution and downscaling does help, consistent with our diagnosis, since the higher-resolution pass lifts many boxes above the token-cell scale. The gain is modest for its cost, however. It buys 3.2 points of $\mathrm{Det}_4$ at four times the tokens and correspondingly slower generation, degrades Appearance well beyond seed noise (0.886 to 0.902), and still trails our three variants.

\smallbreak\noindent\textbf{CLAG.} The two attention gains alone, with no change to the weights, raise Detectability by 8.0 points over the baseline. The improvement is spread over all four axes, as expected of an edit that lets the image read the whole condition stream more. Appearance does not degrade at the selected setting.
% ; on this checkpoint a moderate condition gain slightly improves it (Section~\ref{subsec:ablations}).

\smallbreak\noindent\textbf{Layout-Weighted Loss.} Using the Layout-Weighted Loss ($W{=}3$) raises Detectability by 9.0, improving presence and class the most. 
% This is the pattern the diagnosis predicts. 
The reweighted objective gives the
model more incentive to produce the object that a covered location calls for, namely its existence and its identity. The weight is chosen from a sweep in the supplementary, where independent runs at high $W$ reproduce the gain, so the improvement is not an artifact of a single training run.
% — but no new information about where the box boundary lies, which is fixed by the layout at inference regardless of the weight.

\smallbreak\noindent\textbf{Combination.} Applying the attention gains on top of the reweighted checkpoint gives the best Detectability in the table, 85.34 ± 0.66 over five seeds. The two families are complementary in the axes they move. Training brings presence and class down, and the inference-time gains then bring localization and size down as well. 
% The gain that the reweighted model needs is smaller than what the baseline needs — x=1.2 against x=1.5 — which Section~\ref{subsec:ablations} examines. 
Appearance is unchanged within seed noise (0.887±0.002 for both the reweighted checkpoint and the combination), so at the selected setting the attention gains buy Detectability at no measurable Appearance cost.
The finer trend, and its attribution to the condition gain rather than the sub-token gain, are in Section~\ref{subsec:ablations}.
These gains do not come from drawing extra objects, as detections outside the layout fall from 10.1 per image for the baseline to 5.8 for the combination, against 4.7 on real images.

\smallbreak\noindent\textbf{Scene-level metrics.} CLIPScore and PickScore move little across all rows. Both are computed at 224 pixels, where an object of 8–16 pixels in the original image is not resolved, so they measure the scene rather than the objects the layout asks for. We report them to show that the edits do not damage the scene, not as evidence of tiny-object fidelity; that is the gap TOGBench is built to fill.

% We now evaluate the two methods on TOGBench. The baseline is already strong on presence, class, and appearance, so an improvement must appear on the size and shape axes without regressing the others. We compare the baseline generator, each method, and their combination, and we additionally include two training-free baselines, condition-decoupled guidance and two-times generation with downscaling \cite{}. We report training and inference cost so that the retraining method and the training-free methods are compared fairly.

% TODO: T1 main comparison table. Rows: Real GT / Baseline / +M1 / +M2 / +M1+M2 / condition-CFG / 2x-gen.
% Columns: overall Fidelity + 8 per-axis severities (size, shape highlighted) + Airplane Fidelity + train GPU-h + inference cost.
% Only Real (100.0) and Baseline (68.2, airplane 66.5) are known; all other cells PENDING.
% TODO: T4 per-class Fidelity, baseline column known (storage-tank 95.3, swim-pool 93.7, ship 93.3, vehicle 90.5,
%       person 90.5, airplane 66.5, wind-mill 27.3 low-N, bridge UNSCORED); method columns PENDING.
% TODO: F3 per-axis lift bars (baseline vs +M1 vs +M2 vs +both), size and shape emphasized.
% TODO: F5 qualitative crop grid, folded in here: rows airplane/ship/storage-tank, columns Real/Baseline/+M1/+M2/+both,
%       at the 2x marked-crop resolution the judge sees, plus one honest failure row. Shows FIXES (distinct from Fig. in 5.5, which shows failures).

\subsection{Analysis and Ablations}
\label{subsec:ablations}
% SCAFFOLD -- PENDING method runs. Structure and captions fixed; numbers to drop in.

%We ablate the components of each method and analyze why the training-free method works.

We ablate the three components of Section~\ref{sec:method} and test the two interaction predictions of Section~\ref{subsec:threeinteract}. Unless stated otherwise, every number below is from a single generation seed on the fixed 500 layouts; Table~\ref{tab:tog-main} reports the selected configurations over five seeds.

\smallbreak\noindent\textbf{The two gains share a budget.} Table~\ref{tab:grid} sweeps the condition gain $x$ and the sub-token cell gain $y$ over a $5{\times}5$ grid, on the baseline checkpoint and on the reweighted one. Both grids show the same qualitative pattern. Along $x$ alone ($y{=}1$), $\mathrm{Det}_4$ rises and then falls; along $y$ alone ($x{=}1$), it rises monotonically over the range tested, at a small cost in Appearance. Per point of $\mathrm{Det}_4$ gained, the sub-token gain moves Appearance by about a fifth of what the condition gain does, consistent with its touching under one percent of the condition keys (Figure~\ref{fig:method}c). 
% Their benefits do not add. 
In every row the best $y$ falls as $x$ rises, from $y{=}2.0$ at $x{=}1.0$ to $y{=}1.0$ at $x{=}2.0$ on both checkpoints, which is what Section~\ref{subsec:threeinteract} predicts when the sub-token cells are amplified by the product $xy$. The best cell lies on this ridge in both grids.

\smallbreak\noindent\textbf{Training moves the optimum.} The reweighted checkpoint reaches its best $\mathrm{Det}_4$ at a lower condition gain than the baseline ($x{=}1.5$ against $x{=}1.8$ in Table ~\ref{tab:grid}), and the cells selected for Table~\ref{tab:tog-main} shift by the same one grid step ($x{=}1.2$ against $x{=}1.5$). A model trained to follow the layout needs less of the correction that $x$ supplies, so the $x$ that helps the baseline over-corrects it. Appearance shows the same shift with the opposite sign: on the baseline a moderate $x$ improves Appearance, whereas on the reweighted checkpoint every $x>1$ worsens it.

\smallbreak\noindent\textbf{Selection and seeds.} For Table~\ref{tab:tog-main} we select, on each checkpoint, the knee of the $\mathrm{Det}_4$–Appearance trade-off in Table~\ref{tab:grid} marked by the shaded cell. Beyond it, a further point of $\mathrm{Det}_4$ costs several times more Appearance than any step up to it.
 The selected cell is re-evaluated over five generation seeds. Along y, adjacent cells differ by roughly the seed standard deviation (0.7 $\mathrm{Det}_4$, 0.002 Appearance), so the exact y is not critical; the variation along x, and the shift between checkpoints, are far larger.

% TODO: T5 Method 1 ablation. Sweep gamma (primary), A_0, w_max. Control rows: gamma=0 / w=1 must recover the baseline;
%       w_max=infinity exposes the instability the cap prevents. Report Fidelity, size severity, shape severity, and large-object
%       fidelity as the trade-off control. Add a compute-matched control (uniform small-object upweight) so the gain is the
%       area-inverse reweighting, not extra training budget.
% TODO: T6 Method 2 ablation. Sweep scale s (s=1 is an exact no-op recovering the baseline) and the layer band
%       (peak {15..20} vs early vs late vs all vs none) and band width. Only the peak band should lift Fidelity.
% TODO: F4 output analysis, resolves the Section 6.2 placeholder. Per-layer large-box vs small-box output-magnitude gap
%       peaking at layers ~15--20 (baseline curve = fig:layermag), plus the post-edit small-box curve rising toward the
%       large-box reference within the band, and the generated size-ratio distribution shifting toward 1.0. Tie gap-closure
%       per layer to the size-axis lift.
% The Fidelity score behaves predictably under its aggregation exponent. On the baseline generator the score is $78.0$, $68.2$, and $55.7$ at $p=1$, $2$, and $4$, since a higher exponent penalizes the weakest axes more strongly. The full sweep, and its effect on the ranking of methods, is in the supplementary material.
% TODO: confirm the method ranking is preserved across p once the method runs land.

%% file: sec/7_conclusion.tex
\section{Conclusion}
\label{sec:conclusion}
We introduced TOGBench, a benchmark for the fidelity of tiny-object generation in aerial imagery. It compares generated objects to real objects of the same class along five interpretable axes and reports two scores, Detectability and Appearance. A controlled degradation study confirms that both scores track generation quality. We also proposed two simple techniques that improve TOGBench performance, a loss that reweights small bounding-box regions and an inference-time attention gain on the condition stream. Next steps include extending TOGBench to other domains where fine detail is critical, and correspondence-based axes that match generated objects to real ones with learned local descriptors~\cite{lee2021distill,lee2023rotation} or semantic correspondence~\cite{kim2022transformatcher,min2023chm,kim2024hypercolumn,kim2025sasm}. We expect TOGBench to serve as a measurable objective for faithful tiny-object generation.

%% file: supp.tex
\maketitlesupplementary
% supplementary numbering (S1, S2, ...) so it does not collide with Table 1 / Figure 1 of the main paper
\renewcommand{\thetable}{S\arabic{table}}
\renewcommand{\thefigure}{S\arabic{figure}}
\renewcommand{\theequation}{S\arabic{equation}}

\appendix

\section{Why Tiny Objects Are Hard to Generate}\label{app:challenges}

We give the background moved out of the main paper for space.
The failures TOGBench measures arise from three difficulties intrinsic to the scale of the objects.

\smallbreak\noindent\textbf{Latent-grid limitation.}
Latent diffusion encodes the image into a compressed latent grid before denoising~\cite{rombach2022ldm}.
An object spanning a few pixels may map to a single latent cell, which leaves little capacity for its shape and internal structure.
When an object is under about 8px it can fall below the resolution of even one latent cell, so it has no latent unit of its own and shares a cell with its surroundings.
The model cannot resolve structure that its latent grid does not represent.

\smallbreak\noindent\textbf{Scale magnifies error.}
A few-pixel displacement is minor for a large object but can move a tiny object off its location entirely.
A small absolute size error changes a tiny object's area by a large fraction.
The decoder that maps latents back to pixels further limits fine detail, since it is trained to reconstruct whole scenes rather than to preserve individual small structures~\cite{esser2021vqgan}.

\smallbreak\noindent\textbf{Weak training signal.}
A tiny object covers few pixels, contributes little to the training loss, and tends to be underfit~\cite{cheng2023towards, ye2025objectfidelity}.
The result is objects that are malformed, mis-sized, or missing even when the surrounding scene looks realistic.
This is the imbalance that the Layout-Weighted Loss (Section~5.2) directly targets, by raising the loss weight on the latent locations that layout boxes cover.

\section{Benchmark Details}\label{app:bench}

\smallbreak\noindent\textbf{Composition.}
The benchmark fixes 500 real scenes from the AI-TOD-v2 test split~\cite{wang2021aitod,xu2022detecting}, containing 4,459 annotated objects across seven classes.
Per-class counts are given in Table~\ref{tab:composition}.
The objects are tiny: the median object side is about 12px measured as $\sqrt{\text{area}}$, and most objects are between 8 and 22px.
Images are $800\times800$.
AI-TOD provides no text prompts, so one prompt per scene is produced by captioning the real image with the vision-language model Qwen3-VL-32B-Instruct~\cite{bai2025qwen3vl}.
Generation for a scene uses that scene's real layout together with this prompt.

\begin{table}[t]
\caption{Per-class object counts over the 500 fixed test scenes.}
\centering\small
\begin{tabular}{lr}
\toprule
Class & Objects \\
\midrule
airplane & 525 \\
bridge & 557 \\
storage-tank & 581 \\
ship & 635 \\
swimming-pool & 249 \\
vehicle & 1{,}266 \\
person & 646 \\
\midrule
Total & 4{,}459 \\
\bottomrule
\end{tabular}
\label{tab:composition}
\end{table}

\smallbreak\noindent\textbf{Instruments and matching.}
Detectability is read out by an ensemble of three detectors, each fine-tuned on AI-TOD-v2: a one-stage detector~\cite{jocher2026yolo26}, a query-based transformer detector~\cite{hu2025domedetr}, and a detector specialized for tiny objects~\cite{huang2024dqdetr}.
Each runs at native $800\times800$ with confidence threshold $\tau=0.3$, and the axes are pooled over the three, which reduces the bias of any single detector.
The full constant list is as follows.
Matching uses a size-scaled radius $\kappa=1$, so a detection matches a box when its center lies within $\kappa\sqrt{\text{box area}}$ of the box center.
Assignment is greedy and nearest-first, and class-agnostic, so placement and class are measured independently.
The per-axis severity is clamped to $[0,\sigma]$ with $\sigma=1.5$.
A matched pair is mis-sized when $\sqrt{\text{area ratio}}>\eta$ or $<1/\eta$, with $\eta=1.3$.
The quality axes (localization, class, and size) require at least $\text{MIN\_N}=50$ matched detections for a class, otherwise that class is skipped for those axes, whereas presence is always kept, because a rate computed on very few matches is noisy.

\begin{figure*}[t]
  \centering
  \includegraphics[width=0.62\linewidth]{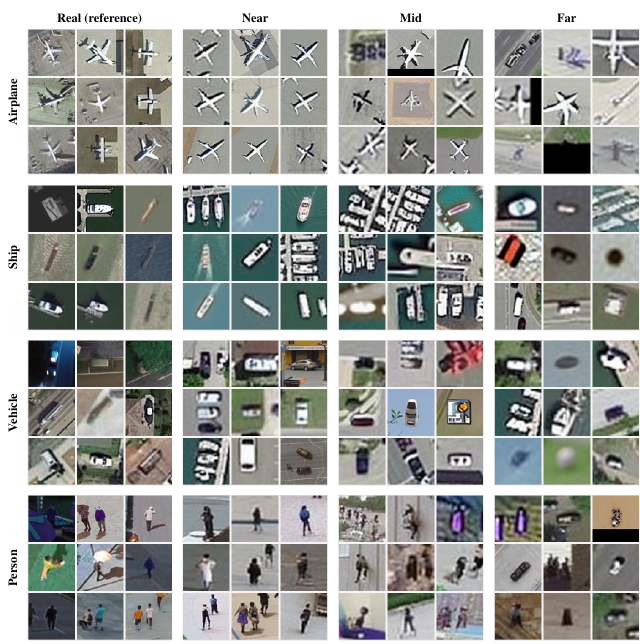}
  \caption{\textbf{Appearance is a set-to-set distance and tracks visual quality.}
  For four classes, the real reference set (left) beside generated sets drawn from three distance bands (near, mid, far).
  Each cell is a set of nine context-padded crops.
  As the distance rises, the generated objects drift from the real distribution: sharp and correctly shaped at small distance, then blurred, malformed, or the wrong object at large distance.}
  \label{fig:appearance}
\end{figure*}

\smallbreak\noindent\textbf{Appearance protocol.}
Each object is cropped with surrounding context (context factor 1.8) and resized to 224px, then embedded with a DINOv3 model pretrained on satellite imagery~\cite{simeoni2025dinov3}.
Features are taken at two scales, 32px and 48px, then L2-normalized and concatenated.
For each generated crop, Appearance is the mean Euclidean distance to its $k=25$ nearest real crops of the same class in a fixed reference bank of 24,933 real crops, as in Equation~(11).
Distances are averaged over crops and then over classes, and lower is better.
We use a nearest-neighbor distance rather than a Fr\'echet distance because a Fr\'echet distance~\cite{heusel2017gans} and its kernel variant~\cite{binkowski2018demystifying} assume many samples and an approximately Gaussian feature distribution, neither of which holds for the few tiny crops available per class~\cite{jayasumana2024rethinking}.
A nearest-neighbor distance, related to precision and recall style measures~\cite{kynkaanniemi2019precision,naeem2020density}, is stable in this small-sample regime.
Figure~\ref{fig:appearance} shows that this distance tracks visual quality: generated sets at increasing distance drift visibly from the real reference set.

\section{Extended Validation}\label{app:validation}

This appendix gives the numeric backing for the validation claim in Section~4.4.

\smallbreak\noindent\textbf{Degradation ladder.}
We take a 100-scene subset of real test scenes, treat their real objects as a known-quality reference, and degrade the scenes with four corruption families at three graded severities each.
The families are JPEG compression, Gaussian blur, Gaussian noise, and object inpainting.
Table~\ref{tab:degrade} reports Detectability ($\mathrm{Det}_4$, higher is better) and Appearance (a distance, lower is better) at each level.
Within every family, as severity rises Detectability falls and the Appearance distance rises, both monotonically, so the two scores track genuine quality loss rather than the choice of instrument.
The two scores emphasize different corruptions: object inpainting removes objects, so it lowers Detectability sharply while barely moving Appearance, whereas blur and noise degrade object texture, so they raise Appearance sharply.
This is why the two scores are reported separately and not merged.
Figure~1 is the visual version of Table~\ref{tab:degrade}.

\begin{table}[t]
\caption{Degradation ladder on a 100-scene subset. Within every family, higher severity lowers Detectability and raises the Appearance distance.}
\centering\small
\begin{tabular}{llcc}
\toprule
Family & Severity & $\mathrm{Det}_4\uparrow$ & Appearance $\downarrow$ \\
\midrule
Clean reference & n/a & 100.00 & 0.8175 \\
\midrule
\multirow{3}{*}{JPEG compression}
 & Mild     & 95.51 & 0.8513 \\
 & Moderate & 91.60 & 0.8646 \\
 & Severe   & 83.40 & 0.8872 \\
\midrule
\multirow{3}{*}{Gaussian blur}
 & Mild     & 93.79 & 0.8892 \\
 & Moderate & 72.82 & 0.9501 \\
 & Severe   & 43.56 & 0.9907 \\
\midrule
\multirow{3}{*}{Gaussian noise}
 & Mild     & 91.31 & 0.9609 \\
 & Moderate & 82.90 & 0.9909 \\
 & Severe   & 62.10 & 1.0118 \\
\midrule
\multirow{3}{*}{Object inpainting}
 & Mild     & 95.54 & 0.8373 \\
 & Moderate & 78.19 & 0.8434 \\
 & Severe   & 55.30 & 0.8584 \\
\bottomrule
\end{tabular}
\label{tab:degrade}
\end{table}

\smallbreak\noindent\textbf{Layout perturbation.}
The ladder validates the scores against image quality; this part validates the four detection axes against the layout that conditions generation.
On a 100-scene set we corrupt one property of the layout at a time, regenerate, and re-score against the original ground truth.
Each axis is reported as a real-anchored severity, where 0 is as good as real and higher is worse.
Table~\ref{tab:perturb} gives the result.
Dropping boxes raises presence and leaves the other axes near baseline.
Rescaling boxes raises size, and localization with it, since a resized object also overlaps its box less.
Relabeling classes raises class severity monotonically with dose (0.04 to 0.06 from 30 to 100 percent relabeled), though modestly: at a few pixels an object carries little class-discriminative appearance, so a relabeled object is often still detected and classified from context.
The class axis itself is discriminative on genuine generation errors, where it separates methods in Table~1; here it is the relabel perturbation, not the axis, that is weak.
A seed-only control, which changes the generation randomness but not the layout, stays at baseline.
Each axis therefore responds to the corruption it targets, and an irrelevant change moves nothing.

\begin{table}[t]
\caption{Layout perturbation on a 100-scene set. Each entry is a real-anchored severity (0 is real-like, higher is worse), pooled over the seven classes and three detectors. Corrupting one layout property raises the axis it targets, presence for dropping, size and localization for rescaling, and class for relabeling, while a seed-only control stays at baseline. Doses are representative and the response is monotone in dose.}
\centering\small
\begin{tabular}{lcccc}
\toprule
Perturbation & Presence & Localization & Class & Size \\
\midrule
Baseline (generation)       & 0.06 & 0.12 & 0.04 & 0.09 \\
Seed control                & 0.07 & 0.13 & 0.03 & 0.08 \\
\midrule
Drop 40\% of boxes          & 0.36 & 0.14 & 0.04 & 0.10 \\
Relabel all classes         & 0.11 & 0.12 & 0.06 & 0.08 \\
Rescale boxes ($\times 1.6$) & 0.21 & 0.45 & 0.03 & 0.62 \\
\bottomrule
\end{tabular}
\label{tab:perturb}
\end{table}

\section{Method Details and Analysis}\label{app:method}

\smallbreak\noindent\textbf{Choice of the reweighting weight $W$.}
Table~\ref{tab:wsweep} sweeps the reweighting weight with three independent training runs per value. Training uses no seed, so runs within a row differ only in their stochastic trajectory, while the generation seed is fixed. The spread within a row is therefore training variance alone. This variance is not negligible, reaching $\pm2.9$ $\mathrm{Det}_4$, so we compare weights by run means rather than by single runs. The ladder rises steeply up to $W{=}3$ ($70.6 \rightarrow 74.9 \rightarrow 80.9$) and flattens beyond it ($W{=}10$: $79.5$). This is the pattern Equation~(14) predicts: the box-token share of the objective grows quickly at small $W$ and saturates at large $W$. The gain is carried by the two layout-reading axes, with presence falling from $0.30$ to $0.19$ and class from $0.29$ to $0.09$, while the two geometric axes move less (localization 0.146 to 0.123, size 0.187 to 0.137). This is consistent with the diagnosis of Section~5.1. Reweighting adds incentive to produce the requested object, not new information about where its boundary lies. 
We therefore use $W{=}3$. It is the smallest weight whose mean gain clears the
run-to-run spread, and among such weights it perturbs the objective least: by Equation~(14) the box-covered share of the loss is $s(3){=}4.2\%$ against $s(10){=}12.8\%$. We do not claim $W{=}3$ is better than $W{=}10$, their means differ by 1.4~$\mathrm{Det}_4$ against a spread of $\pm0.9$--$1.2$ only that it is the cheaper of the two. All inference-time comparisons (Section~6.3) hold this single checkpoint fixed, so they do not depend on which run is drawn.

\begin{table*}[t]
\centering\small
\setlength{\tabcolsep}{4pt}
\caption{Loss-reweighting weight sweep. $\mathrm{Det}_4$ (higher is better) and Appearance (lower is better) as mean $\pm$ sd over three independent training runs per weight; $W{=}1$ is the unweighted objective of Equation~(2), retrained under the same protocol. Scores rise steeply up to $W{=}3$ and flatten beyond it, with a
run-to-run spread that leaves the top two weights indistinguishable.}
\label{tab:wsweep}
\newcommand{\sdx}[1]{{\scriptsize$\pm$#1}}
\begin{tabular}{@{}lcccc cc@{}}
\toprule
$W$ & Pres.$\downarrow$ & Loc.$\downarrow$ & Class$\downarrow$ & Size$\downarrow$ & $\mathrm{Det}_4\uparrow$ & App.$\downarrow$ \\
% \begin{tabular}{@{}lcc@{}}
% $W$ & Det$_4\uparrow$ & App.$\downarrow$ \\
\midrule
  % 1 (baseline)  & 70.63\,\sdx{2.87} & 0.8947\,\sdx{0.0057} \\
  % 1.5           & 74.85\,\sdx{1.82} & 0.8883\,\sdx{0.0036} \\
  % 2           & 0.000\,\sdx{9.999} & 0.000\,\sdx{9.999} \\
  % 3             & \textbf{80.93}\,\sdx{0.93} & 0.8848\,\sdx{0.0035} \\
  % 5           & 0.000\,\sdx{9.999} & 0.000\,\sdx{0.000} \\
  % 10            & 79.51\,\sdx{1.18} & 0.8877\,\sdx{0.0006} \\
  1 (baseline) & 0.302\,\sdx{0.087} & 0.146\,\sdx{0.017} & 0.288\,\sdx{0.134} & 0.187\,\sdx{0.025} & 70.63\,\sdx{2.87} & 0.8947\,\sdx{0.0057} \\
  1.5          & 0.243\,\sdx{0.020} & 0.142\,\sdx{0.014} & 0.164\,\sdx{0.026} & 0.170\,\sdx{0.019} & 74.85\,\sdx{1.82} & 0.8883\,\sdx{0.0036} \\
  % 2 & 0.248\,\sdx{0.024} & 0.157\,\sdx{0.003} & 0.159\,\sdx{0.025} & 0.187\,\sdx{0.007} & 73.80\,\sdx{2.33} & 0.8951\,\sdx{0.0042} \\
  3            & 0.187\,\sdx{0.005} & 0.123\,\sdx{0.013} & 0.090\,\sdx{0.023} & 0.137\,\sdx{0.009} & \textbf{80.93}\,\sdx{0.93} & 0.8848\,\sdx{0.0035} \\
  % 5 & 0.252\,\sdx{0.088} & 0.132\,\sdx{0.034} & 0.152\,\sdx{0.044} & 0.136\,\sdx{0.048} & 79.50\,\sdx{1.49} & 0.8907\,\sdx{0.0007} \\
  10           & 0.195\,\sdx{0.013} & 0.123\,\sdx{0.015} & 0.115\,\sdx{0.011} & 0.148\,\sdx{0.017} & 79.51\,\sdx{1.18} & 0.8877\,\sdx{0.0006} \\
\bottomrule
\end{tabular}
\end{table*}

\smallbreak\noindent\textbf{The condition gain is an odds multiplier.}
The CLAG of Section~5.3 adds $\log(x)$ to the condition logits before the softmax of Equation~(15). Let $m$ be the share of an image query's attention mass that falls on the condition stream. Adding $\log x$ multiplies the exponentiated logits of the condition keys by $x$ and leaves the others unchanged, so after renormalization the odds of attending to the condition, $m/(1-m)$, are multiplied by exactly $x$, while $m$ itself saturates and grows by less. The odds, not the mass, is what the gain controls, so $x$ is
a stated quantity rather than an opaque tuned knob. To check that this survives contact with the full model, we instrument the joint attention on the baseline checkpoint and recompute the softmax explicitly, recording where each image query places its mass, averaged over heads, queries, the 19 dual-stream and 38 single-stream blocks, and three denoising steps (the first, middle, and last of 28), over all $500$ benchmark layouts and with the sub-token gain off ($y{=}1$). The shares move little across the sampled steps and the per-block profile is stable across layouts.
Table~\ref{tab:odds} reports the mass on the condition stream, the measured odds ratio relative to $x{=}1$, and the odds ratio predicted by the equation.

\begin{table}[t]
\centering\small
\begin{tabular}{@{}lccccc@{}}
\toprule
$x$ & Text & Image & Cond. & Measured odds & Predicted \\
\midrule
  1   & 0.2122 & 0.7223 & 0.0655 & 1.000 & 1.0 \\
  1.2 & 0.2101 & 0.7132 & 0.0767 & 1.185 & 1.2 \\
  1.5 & 0.2078 & 0.6996 & 0.0926 & 1.455 & 1.5 \\
  2   & 0.2046 & 0.6782 & 0.1171 & 1.892 & 2.0 \\
\bottomrule
\end{tabular}
\caption{The condition gain $x$ as an odds multiplier. Image-query mass on the condition stream, the measured odds ratio, and the odds ratio predicted by Equation~(15).}
\label{tab:odds}
\end{table}

The read-out shows three things. First, the imbalance the gain is meant to correct: at $x{=}1$ there are equally many image and condition keys (2,500 each, against 512 text keys), yet image queries put 0.722 of their mass on other image tokens and only 0.066 on the condition, an 11$\times$ gap that widens with depth (about 0.127 in the dual-stream blocks against 0.047 in the single-stream blocks).
Second, the gain does what the equation says: the measured odds ratios (1.185, 1.455, 1.892) track the predicted $x$ at 95 to 99 percent, with a shortfall that grows with $x$ because the condition keys are themselves updated through the same attention, a feedback effect rather than noise. 
The gain $x$ therefore has a concrete meaning, the factor by which the model reads its layout, which is why the optimal $x$ is lower for a reweighted checkpoint whose training has already closed part of the gap (Section~6.3).
Third, the extra mass is reallocated from the image, not from the prompt: between $x{=}1$ and $x{=}2$ the condition gains about 0.052 of mass while the image's own self-attention loses about 0.044 and text only about 0.008. The read-out sweeps $x$, the gain applied to every condition key; $y$ multiplies the odds on the sub-token cells $S$ by the same identity (Section~5.3) and is not measured separately.

\smallbreak\noindent\textbf{Implementation details.}
The generator is FLUX.1-dev~\cite{blackforest2024flux} with the OminiControl spatial-control LoRA adapter~\cite{tan2024ominicontrol}, using 28 denoising steps at $800\times800$, with the layout supplied as a class-colored bounding-box mask.
The Layout-Weighted Loss uses $W{=}3$ at training step 9000.
The CLAG settings carried to Table~1 are the shaded cells of Table~2: on the baseline checkpoint the condition gain $x{=}1.5$ and the sub-token cell gain $y{=}1.2$, and on the reweighted checkpoint $x{=}1.2$ and $y{=}1.8$; the sub-token threshold is the token cell size (16px) and is not tuned.
All comparisons are paired, using the same real layouts and reusing the baseline seed for the inference-time method, and are reported as means over five seeds. The adapter is rank $r{=}4$ with $\alpha{=}4$ and Gaussian initialization, giving 14.5\,M trainable parameters with the FLUX backbone frozen.
We train using the Prodigy optimizer with a learning rate of $1$, weight decay of $0.01$, bias correction, and safeguard warm-up enabled. We use a batch size of 4 without gradient accumulation, gradient checkpointing, and bfloat16 precision at $768\times768$ resolution, and train for 9000 steps on a single NVIDIA A100 80GB GPU.

\section{Additional Qualitative Results}\label{app:qual}

\begin{figure*}[t]
  \centering
  \includegraphics[width=0.9\linewidth]{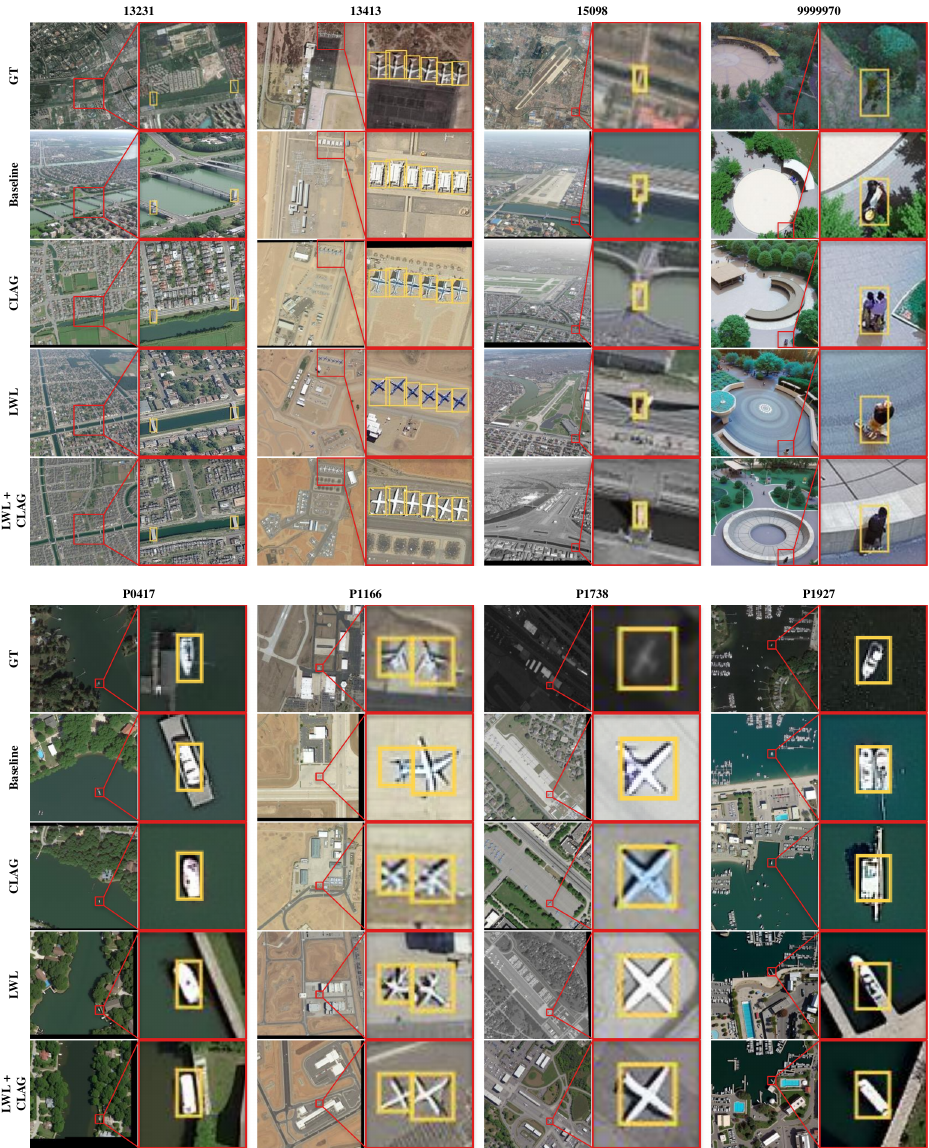}
  % figure to be rendered
  \caption{Per-class before and after comparison on benchmark layouts.
  Rows are the real image (GT), Baseline, CLAG, Layout-Weighted Loss (LWL), and LWL + CLAG, and columns are eight scenes, each shown in full with a magnified crop of the target tiny objects (yellow boxes).
  The baseline omits objects or renders the wrong class at requested box locations; the reweighted model makes those objects present and correctly classed; CLAG then corrects placement and size and brings back sub-token objects; the combination is the most consistent across classes. Column headers refer to actual filenames of the evaluation set.}
  \label{fig:supp_perclass}
\end{figure*}

Figure~\ref{fig:supp_perclass} shows per-class before and after comparisons of the baseline generator and our two techniques on benchmark layouts, complementing the main qualitative figures (Figures 4 and 5).
The trends match the axis story of Section~6.2.

Presence and class improve with the Layout-Weighted Loss.
Objects that the baseline drops entirely, or paints with the color of a different class, become present and correctly classed once box-covered locations receive more loss weight.
Localization and size improve with the attention gains, which pull generated instances toward their requested box centers and correct boxes rendered too large or too small.
The sub-token cell gain specifically recovers objects whose box longer side is under one token cell, which the baseline tends to omit.
Applying both techniques compounds these effects, consistent with the combined $\mathrm{Det}_4$ of 85.34 against the baseline 72.19 in Table~1, though a residual gap to the Real column remains in the finest object detail.

% \begin{figure*}[t]
%   \centering
%   \includegraphics[width=\linewidth]{figures/supp_failures}
%   % figure to be rendered
%   \caption{Failure cases that remain after applying both techniques: very dense clusters where neighboring instances merge, objects under about 8px that stay below the latent-cell resolution and lose internal structure (Appendix~\ref{app:challenges}), and confusions between visually similar classes.}
%   \label{fig:supp_failures}
% \end{figure*}

\section{Additional Benchmark Analyses}\label{app:cr}

This appendix reports analyses of the benchmark itself: how much the scores depend on the detectors, per-class results, whether the methods produce objects the layout does not request, and the object-size distribution.
All numbers are computed from the same detector predictions and score files as Table~1, with the scorer unchanged.

\smallbreak\noindent\textbf{Dependence on the detector.}
A detector may behave differently on real and generated images, so we re-score every method with each detector alone, using the same detector for the real anchor (Table~\ref{tab:cr-detectors}).
Absolute scores differ between detectors, but the ranking of the methods is the same under every detector.

\begin{table}[t]
\caption{$\mathrm{Det}_4$ (higher is better) under each detector alone and under the ensemble, with the real anchor scored by the same detector(s). Mean $\pm$ sd over five generation seeds. $^\ddagger$Seed 42 only, since the seed sd of $\mathrm{Det}_4$ (at most 1.0 in Table~1) is small next to the gaps between these rows and the others.}
\label{tab:cr-detectors}
\centering\footnotesize
\setlength{\tabcolsep}{3pt}
\resizebox{\linewidth}{!}{%
\begin{tabular}{@{}lcccc@{}}
\toprule
Method & Ensemble & YOLO26x & Dome-DETR & DQ-DETR \\
\midrule
Guidance scale $=1.0$$^\ddagger$ & 60.03 & 31.81 & 66.51 & 42.24 \\
Guidance scale $=3.5$ (default) & 72.19\,{\scriptsize$\pm$0.27} & 62.02\,{\scriptsize$\pm$0.91} & 72.85\,{\scriptsize$\pm$0.53} & 68.91\,{\scriptsize$\pm$4.37} \\
Guidance scale $=6.0$$^\ddagger$ & 68.87 & 61.75 & 69.84 & 68.79 \\
$2\times$ inference resolution$^\ddagger$ & 75.89 & 64.58 & 77.94 & 75.79 \\
\midrule
Ours (CLAG)             & 80.14\,{\scriptsize$\pm$1.00} & 74.80\,{\scriptsize$\pm$6.49} & 82.18\,{\scriptsize$\pm$0.95} & 78.87\,{\scriptsize$\pm$1.38} \\
Ours (Layout-Weighted Loss) & 81.14\,{\scriptsize$\pm$0.63} & 78.07\,{\scriptsize$\pm$1.05} & 82.99\,{\scriptsize$\pm$0.38} & 80.90\,{\scriptsize$\pm$0.83} \\
Ours (CLAG + Layout-Weighted Loss) & 85.34\,{\scriptsize$\pm$0.66} & 81.25\,{\scriptsize$\pm$1.41} & 87.50\,{\scriptsize$\pm$0.62} & 84.99\,{\scriptsize$\pm$0.86} \\
\bottomrule
\end{tabular}}
\end{table}

\smallbreak\noindent\textbf{Per-class results.}
Table~\ref{tab:cr-perclass} breaks the scores of Table~1 down by class.
Bridge is the hardest class for every method and gains the most ($36.5$ to $75.1$ with both techniques), and CLAG improves all seven classes.
The bridge deficit is not a detector artifact, as the three detectors find real bridges at a recall of $0.70$.

\begin{table*}[t]
\caption{Per-class results for the methods of Table~1. Top: per-class Detectability $D_c$ (Equation~(9), higher is better), mean $\pm$ sd over five seeds. Bottom: per-class Appearance distance (lower is better), mean over five seeds.}
\label{tab:cr-perclass}
\centering\footnotesize
\setlength{\tabcolsep}{4pt}
\begin{tabular}{@{}lccccccc@{}}
\toprule
 & airplane & bridge & storage-tank & ship & swimming-pool & vehicle & person \\
\midrule
\multicolumn{8}{@{}l}{\emph{Detectability $D_c$} $\uparrow$} \\
Baseline (guidance 3.5) & 81.35\,{\scriptsize$\pm$2.26} & 36.49\,{\scriptsize$\pm$1.55} & 82.06\,{\scriptsize$\pm$1.85} & 87.54\,{\scriptsize$\pm$3.46} & 86.65\,{\scriptsize$\pm$1.84} & 90.82\,{\scriptsize$\pm$0.76} & 83.60\,{\scriptsize$\pm$0.87} \\
CLAG & 85.57\,{\scriptsize$\pm$2.52} & 55.98\,{\scriptsize$\pm$1.51} & 87.75\,{\scriptsize$\pm$1.02} & 87.89\,{\scriptsize$\pm$1.80} & 89.29\,{\scriptsize$\pm$2.52} & 94.96\,{\scriptsize$\pm$0.55} & 87.00\,{\scriptsize$\pm$0.90} \\
Layout-Weighted Loss & 86.14\,{\scriptsize$\pm$2.19} & 60.89\,{\scriptsize$\pm$1.42} & 92.08\,{\scriptsize$\pm$1.51} & 87.63\,{\scriptsize$\pm$3.53} & 85.22\,{\scriptsize$\pm$0.84} & 92.59\,{\scriptsize$\pm$0.50} & 83.78\,{\scriptsize$\pm$0.90} \\
CLAG + Layout-Weighted Loss & 87.91\,{\scriptsize$\pm$1.49} & 75.12\,{\scriptsize$\pm$1.54} & 97.87\,{\scriptsize$\pm$0.21} & 84.13\,{\scriptsize$\pm$4.73} & 85.55\,{\scriptsize$\pm$1.04} & 95.11\,{\scriptsize$\pm$0.86} & 84.88\,{\scriptsize$\pm$0.87} \\
\midrule
\multicolumn{8}{@{}l}{\emph{Appearance} $\downarrow$} \\
Baseline (guidance 3.5) & 0.894 & 0.969 & 0.891 & 0.824 & 0.929 & 0.851 & 0.844 \\
CLAG & 0.891 & 0.964 & 0.894 & 0.819 & 0.930 & 0.852 & 0.842 \\
Layout-Weighted Loss & 0.915 & 0.959 & 0.884 & 0.828 & 0.918 & 0.853 & 0.852 \\
CLAG + Layout-Weighted Loss & 0.914 & 0.952 & 0.883 & 0.844 & 0.920 & 0.851 & 0.844 \\
\bottomrule
\end{tabular}
\end{table*}

\smallbreak\noindent\textbf{Objects outside the layout.}
The detection axes are computed over layout boxes, so we count over-generation separately as the detections with confidence at least $\tau$ that match no layout box (Table~\ref{tab:cr-overgen}).
Real images also yield such detections, from unannotated objects and detector false positives, and serve as the anchor.
Each of our variants produces fewer of them than the baseline, while generating at twice the resolution adds them.

\begin{table}[t]
\caption{Over-generation: detections per image (confidence $\geq\tau$) that match no layout box, mean of the three detectors, lower is better. Mean $\pm$ sd over five generation seeds, and $^\ddagger$ marks rows regenerated with generation seed 42 only. Low values also arise when few objects are drawn at all, as at guidance 1.0.}
\label{tab:cr-overgen}
\centering\footnotesize
\begin{tabular}{@{}lc@{}}
\toprule
Method & Over-gen.\,$\downarrow$ \\
\midrule
Real images (anchor) & 4.71 \\
\midrule
FLUX.1-dev (text-only)$^\ddagger$ & 25.04 \\
Guidance scale $=1.0$$^\ddagger$ & 1.88 \\
Guidance scale $=3.5$ (default) & 10.11\,{\scriptsize$\pm$0.84} \\
Guidance scale $=6.0$$^\ddagger$ & 10.96 \\
$2\times$ inference resolution$^\ddagger$ & 15.71 \\
\midrule
Ours (CLAG) & 5.74\,{\scriptsize$\pm$0.26} \\
Ours (Layout-Weighted Loss) & 9.22\,{\scriptsize$\pm$1.74} \\
Ours (CLAG + Layout-Weighted Loss) & 5.77\,{\scriptsize$\pm$0.78} \\
\bottomrule
\end{tabular}
\end{table}

\begin{figure}[t]
  \centering
  \includegraphics[width=\linewidth]{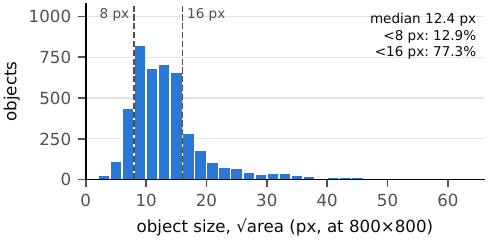}
  \caption{Object sizes in the 500 benchmark scenes, measured as $\sqrt{\text{area}}$ at $800{\times}800$. Dashed lines mark the latent cell (8\,px) and the condition token cell (16\,px) of the generator.}
  \label{fig:cr-sizes}
\end{figure}

\smallbreak\noindent\textbf{Object sizes.}
Figure~\ref{fig:cr-sizes} shows the size distribution of the 4,459 benchmark objects.
The median is 12.4\,px, 12.9\% of the objects are under 8\,px, the size of one latent cell, and 77.3\% are under 16\,px, the size of one condition token cell.
Vehicle is the smallest class (median 10.1\,px, 91\% under 16\,px) and swimming-pool the largest (median 18.5\,px, 37\% under 16\,px).
The scenes come from three AI-TOD-v2 sources: 201 from DOTA and 200 from other satellite and high-altitude aerial sources, both near-nadir, and 99 (19.8\%) from VisDrone, which are low-altitude drone images, often oblique, and contain all 646 person objects and 35 vehicles.

\section{Limitations}\label{app:limitations}
TOGBench has so far been demonstrated with one layout-conditioned generator on AI-TOD-v2, whose scenes are mostly near-nadir aerial views.
Validating it against human judgments and downstream detection, and extending it to other generators, domains, viewpoints, and degradations, is left to future work.
Our techniques improve tiny-object fidelity but do not close the gap to real objects.
The appearance of the very smallest objects is still limited by the latent resolution, since a sub-token object has fewer than one condition token to describe it.